# PyVBMC: Efficient Bayesian inference in Python

**Bobby Huggins** ¹*¶, **Chengkun Li** ¹*, **Marlon Tobaben** ¹*, **Mikko J. Aarnos**¹, and **Luigi Acerbi** ¹¶

**1** University of Helsinki ¶ Corresponding author * These authors contributed equally.

## Summary

PyVBMC is a Python implementation of the Variational Bayesian Monte Carlo (VBMC) algorithm for posterior and model inference for *black-box* computational models (Acerbi, 2018, 2020). VBMC is an approximate inference method designed for efficient parameter estimation and model assessment when model evaluations are mildly-to-very expensive (e.g., a second or more) and/or noisy. Specifically, VBMC computes:

- a flexible (non-Gaussian) approximate posterior distribution of the model parameters, from which statistics and posterior samples can be easily extracted;
- an approximation of the model evidence or marginal likelihood, a metric used for Bayesian model selection.

PyVBMC can be applied to any computational or statistical model with up to roughly 10-15 continuous parameters, with the only requirement that the user can provide a Python function that computes the target log likelihood of the model, or an approximation thereof (e.g., an estimate of the likelihood obtained via simulation or Monte Carlo methods). PyVBMC is particularly effective when the model takes more than about a second per evaluation, with dramatic speed-ups of 1-2 orders of magnitude when compared to traditional approximate inference methods.

Extensive benchmarks on both artificial test problems and a large number of real models from the computational sciences, particularly computational and cognitive neuroscience, show that VBMC generally — and often vastly — outperforms alternative methods for sample-efficient Bayesian inference, and is applicable to both exact and simulator-based models (Acerbi, 2018, 2019, 2020). PyVBMC brings this state-of-the-art inference algorithm to Python, along with an easy-to-use Pythonic interface for running the algorithm and manipulating and visualizing its results.

## Statement of need

Standard methods for Bayesian inference over arbitrary statistical and computational models, such as Markov Chain Monte Carlo (MCMC) methods, traditional variational inference, and nested sampling, require a large number of evaluations of the target density, and/or a model which is differentiable with respect to its parameters (Martin et al., 2020; Murphy, 2023). PyVBMC targets problems that are not amenable to these existing approaches: It uses no parameter gradients, so it is applicable to black-box models, and it is *sample-efficient*, requiring very few likelihood evaluations, typically on the order of a few hundred, as opposed to the many (tens of) thousands required by most other approximate inference methods.

## Method

PyVBMC achieves practical sample efficiency in a unique manner by simultaneously building two approximations of the true, expensive target posterior distribution:

- A Gaussian process (GP) surrogate of the target (unnormalized) log posterior density.
- A variational approximation — an expressive mixture of Gaussian distributions with an arbitrary number of components — fit to the GP surrogate.

The uncertainty-aware GP surrogate is built iteratively via active sampling, selecting which points to evaluate next so as to explore the posterior landscape and reduce uncertainty in the approximation (Figure 1). In this respect, PyVBMC works similarly to Bayesian optimization methods (Garnett, 2023), although with the different objective of learning the full shape of the target posterior as opposed to only searching for the optimum of the target.

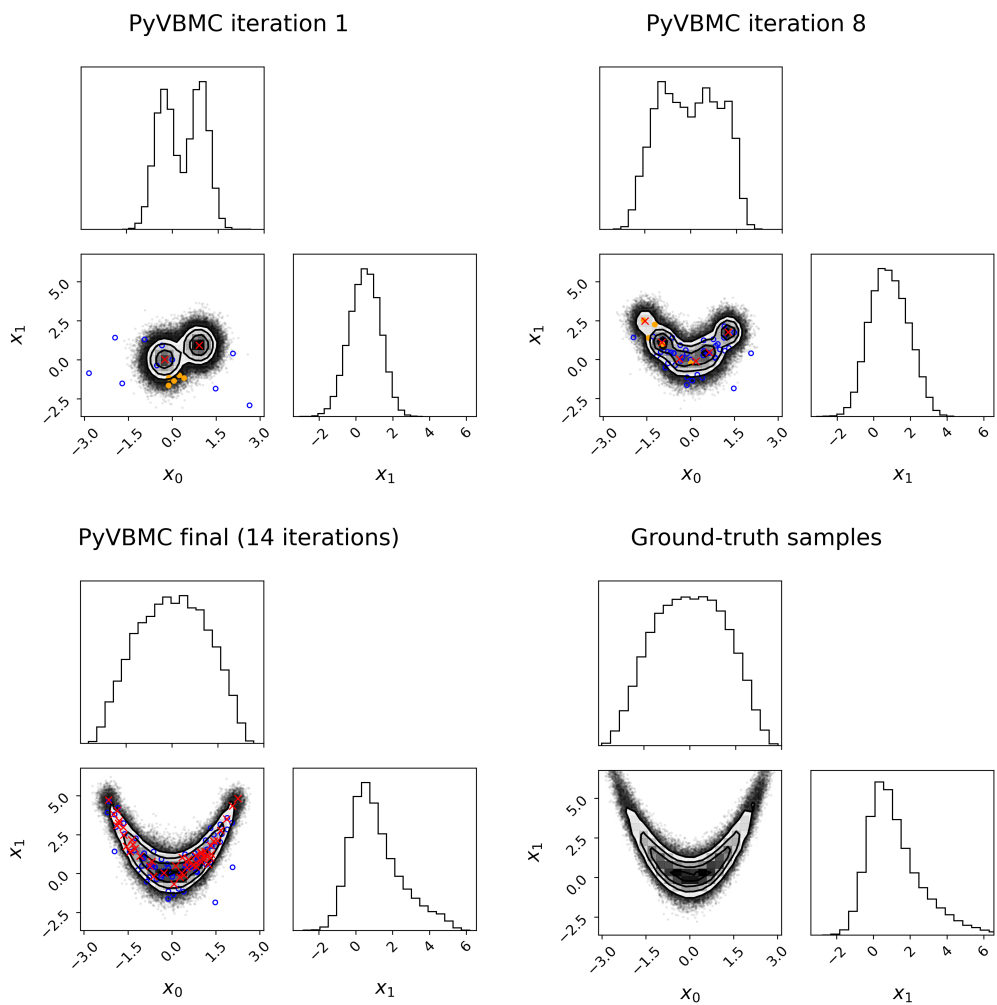

**Figure 1:** Contour plots and marginal histograms show PyVBMC exploring a two-dimensional posterior (a Rosenbrock likelihood with Gaussian prior). The solid orange circles indicate new points chosen by active sampling, the empty blue circles indicate previously sampled points, and the red crosses indicate centers of the variational posterior mixture components. 10 points are sampled initially, and each iteration consists of 5 actively sampled points, meaning that the final result is obtained with only 80 evaluations of the target density.

At the same time, in each iteration of PyVBMC a variational approximation is fit to the current GP surrogate by optimizing a lower bound to the log model evidence (ELBO). This second approximation thus provides an estimate of the normalization constant of the posterior, useful for Bayesian model selection, and yields a tractable distribution (a very flexible mixture of Gaussians) we can easily compute with and draw samples from. Crucially, obtaining this variational approximation is inexpensive because the ELBO and its gradients can be efficiently estimated via Bayesian quadrature (Ghahramani & Rasmussen, 2002; O'Hagan, 1991), without relying on additional evaluations of the true target posterior.

The variational approximation is a unique feature of the VBMC algorithm that makes it particularly efficient and robust, as both the current ELBO and the variational posterior are used throughout all steps of the algorithm, for example:

- as diagnostics for convergence and stability of the solution;
- to obtain a better representation of the target by rotating the axes via *variational whitening*;
- to estimate complex integrated acquisition functions for active sampling with noisy targets.

All of these algorithmic steps would be cumbersome if not infeasible without an easy, tractable representation of the posterior at each iteration. Notably, the GP itself is not tractable as its normalization constant is unknown and we cannot directly draw posterior samples from the surrogate log density: this is where the variational posterior approximation steps in.

In practice, PyVBMC returns the final variational posterior as an object that can be easily manipulated by the user, and an estimate of the ELBO and of its uncertainty. Importantly, several diagnostics are automatically applied to detect lack of convergence of the method.

### Related work

Algorithms predating VBMC, such as WSABI (Gunter et al., 2014), BBQ (Osborne et al., 2012), BAPE (Kandasamy et al., 2015), and AGP (Wang & Li, 2018), employ similar strategies but underperform VBMC in previous experiments, and to our knowledge are not readily available as packaged and user-friendly software for Bayesian inference[1]. The Python package GPry (Gammal et al., 2022), released after VBMC but before PyVBMC, likewise relies on surrogate modeling and active sampling, but learns only a GP surrogate — not a variational posterior — and so generating samples or estimating the model evidence requires further post-processing.

Furthermore, neither GPry nor the aforementioned algorithms are designed to handle noisy log density evaluations. Standard approaches to active sampling based on a surrogate GP can be easily fooled if the model's likelihood is not a deterministic function of the parameters, but is instead measured only up to random error (Järvenpää et al., 2021). Noisy evaluations of the log density occur frequently in simulator-based inference, where the likelihood is estimated from simulated data and statistics thereof. PyVBMC includes separate defaults and robust integrated acquisition functions for working with noisy likelihoods and therefore is particularly well suited to noisy and simulator-based contexts.

### Applications and usage

The VBMC algorithm, in its MATLAB implementation, has already been applied to fields as diverse as neuroscience (Stine et al., 2020), nuclear engineering (Che et al., 2021), geophysical inverse problems (Hao et al., 2022), and cancer research (Demetriades et al., 2022). With PyVBMC, we bring the same sample-efficient and robust inference to the wider open-source Python community, while improving the interface, test coverage, and documentation.

The package is available on both PyPI (`pip install pyvbmc`) and `conda-forge`, and provides an idiomatic and accessible interface, only depending on standard, widely available scientific

---

[1]Emukit (Paleyes et al., 2019) implements WSABI, but targets Bayesian optimization and general-purpose quadrature over inference.

Python packages (Foreman-Mackey, 2016; Harris et al., 2020). The user only needs to give a few basic details about their model and its parameter space, and PyVBMC handles the rest of the inference process. PyVBMC runs out of the box without tunable parameters and includes automatic handling of bounded variables, robust termination conditions, and sensible default settings. At the same time, experienced users can easily supply their own options. We have extensively tested the algorithm and implementation details for correctness and performance. We provide detailed tutorials, so that PyVBMC may be accessible to those not already familiar with approximate Bayesian inference, and our comprehensive documentation will aid not only new users but future contributors as well.

## Acknowledgments

Work on the PyVBMC package is supported by the Academy of Finland Flagship programme: Finnish Center for Artificial Intelligence FCAI.